\documentclass{article}

\usepackage{arxiv}

\usepackage[utf8]{inputenc} % allow utf-8 input
\usepackage[T1]{fontenc}    % use 8-bit T1 fonts
\usepackage{hyperref}       % hyperlinks
\usepackage{url}            % simple URL typesetting
\usepackage{booktabs}       % professional-quality tables
\usepackage{amsfonts}       % blackboard math symbols
\usepackage{nicefrac}       % compact symbols for 1/2, etc.
\usepackage{microtype}      % microtypography
\usepackage{amsmath}        % for \text in math mode
\usepackage{lipsum}
\usepackage{graphicx}
\graphicspath{ {./images/} }
\usepackage{xcolor}

\begin{document}

%\title{Breaking the Hivemind: Meta-Persona Anchoring and Filtered Temperature Scaling to Escape Model Homogeneity}
%\title{Beyond LLM Homogeneity: Escaping the Spirit of the Beehive via Meta-Persona Anchoring and Filtered Temperature Scaling}
\title{Beyond the Hivemind: Escaping LLM Homogeneity via Meta-Persona Anchoring and Sequential Temperature Scaling}

%% --- AUTHORS ---

\author{
  Tairan Fu \\
  Politecnico di Milano \\
  Milano, Italy \\
  \And
  Javier Conde, Carlos Arriaga, Gonzalo Martínez, Pedro Reviriego \\
  Information Processing and Telecommunications Center \\
  Universidad Politécnica de Madrid \\
  Madrid, Spain \\
  \And
  Javier Coronado-Blázquez \\
  Banco de España \\
  Madrid, Spain \\
}
  
\maketitle
\begin{abstract}

Recent studies have identified an ``Artificial Hivemind'' effect in Large Language Models (LLMs) causing models to converge on a narrow, homogenized consensus even for open questions. This semantic collapse limits the diversity of AI, resulting in high inter-response similarity ($\approx 0.80-0.90$) even under high-temperature sampling. In this paper, we propose a novel mitigation framework to increase diversity: Meta-Persona Anchoring combined with Filtered Temperature Scaling (FTS). 

Our approach utilizes a two-stage generation process: first, the model is prompted to self-select a unique, idiosyncratic persona to anchor its starting point; second, we apply a dual-stage sampling sieve, utilizing Top-$p$ filtering to preserve grammatical validity followed by extreme temperature scaling ($T \ge 4.0$) on the surviving candidates to explore the broadened probability distribution. 

We evaluate our method using the INFINITY-CHAT dataset on state-of-the-art open weight models under $\sim$20B parameters. Our results demonstrate a significant reduction in semantic convergence, with average pairwise cosine similarity dropping from ($\approx 0.85$) to ($\approx 0.65$). Our scheme achieves a majority of questions below the 0.7 threshold, effectively reducing the gap between artificial mode collapse and human-level typological diversity. We provide our implementation as an open-source framework to enable more diverse and creative AI deployments.

\end{abstract}

% keywords can be removed

\section{Introduction}
\label{sec:introduction}

The evolution of Large Language Models (LLMs) from foundational Transformer architectures \cite{vaswani2017attention} to sophisticated few-shot learners \cite{brown2020gpt3} has created the need to align the models to human preferences and values. Techniques such as Reinforcement Learning from Human Feedback (RLHF) \cite{ouyang2022instructgpt} have successfully adapted models towards helpfulness and safety. However, this alignment has introduced a systemic side effect: the reduction of generative diversity. 

Early investigations identified a risk of systemic model collapse under recursive training regimes \cite{martinez2023lexical}. This process creates a degradation-prone feedback loop between generative AI and the internet and a baseline for semantic decay. This homogenization is further exacerbated by the mechanisms of instruction tuning; as detailed in recent surveys \cite{zhang2026instruction}, these alignment procedures often prioritize instruction-following consistency at the cost of output variability. This is different from the profound latent diversity found in human conceptual spaces \cite{marti2023latent}, ultimately resulting in a state of pronounced intra- and inter-model homogeneity where the diversity  of the generated text is significantly reduced. This was formalized as the ``Artificial Hivemind'' \cite{jiang2025hivemind}, a state where models across distinct architectural families converge on nearly identical semantic responses to open-ended prompts. Mabrok \cite{mabrok2026latent} provides a topological foundation for understanding this stagnation by formalizing LLM representations as a low-dimensional semantic manifold, suggesting that homogeneity arises when generative trajectories become trapped within high-density regions of this geometric structure.

In this paper, we explore mechanisms to escape the artificial hivemind by modifying token selection and introducing semantic anchoring. By combining Filtered Temperature Scaling (FTS), which decouples token selection from entropy scaling, with Meta-Persona Anchoring, we enable the model to dynamically select its own interpretive lens. This dual-phase approach provides the statistical and semantic permission required to navigate beyond aligned consensus while maintaining logical grounding, demonstrating that the model's idiosyncratic potential remains accessible despite the pressures of alignment.

\section{Background and Related Work}
\label{sec:background}

Recent studies on ``LLM Homogeneity'' and the ``Artificial Hivemind'' \cite{jiang2025hivemind} have identified semantic uniformity in the response of instruction-tuned models to open questions. This phenomenon, often termed mode-collapse, suggests that modern alignment techniques do not merely align models with human values, but steer them into a consensus \cite{zhang2025verbalized}. To mitigate this, "Verbalized Sampling" has been proposed to prompt models to explicitly describe their internal probability distributions as a means of increasing diversity \cite{zhang2025verbalized}. Other works show that the loss of diversity is often localized within the model's final processing stages \cite{zhang2026not}. Most previous efforts to break this homogeneity have generally fallen into two categories: persona-based prompting and stochastic sampling. 

The use of personas by requesting the model to ``speak as a [Role]'' has long been used to elicit stylistic variation. However, research indicates that for highly aligned models, these personas act as a superficial change rather than a true cognitive shift \cite{PersonasLLM}. Studies show that even when prompted with diverse roles, the underlying semantic structure of the response remains tethered to the model's primary alignment prior. This occurs because the model prioritizes the safety and utility constraints of its training over the idiosyncratic requirements of the persona, leading to what has been described as ``different masks on the same face''.

A parallel line of research has attempted to increase diversity through stochastic sampling methods:
\begin{itemize}
    \item \textbf{Temperature Scaling ($T$):} increasing $T$ flattens the distribution, but it does so indiscriminately. At the high-entropy levels required to break consensus, standard sampling begins to include grammatically incoherent tokens, leading to a sharp trade-off between diversity and coherence.
    \item \textbf{Top-$p$ and Top-$k$ Filtering:} prevent the inclusion of ``junk'' tokens but simultaneously reinforce the hivemind by pruning the low probability but valid tokens that lead to idiosyncratic responses. 
    \item \textbf{Min-$p$ Sampling:} filters tokens based on a percentage of the top token's probability, allowing for a more dynamic vocabulary size depending on the model's confidence \cite{minp}. This effectively scales with the distribution's entropy, but it remains an isolated sampling fix and does not change the probabilities of the tokens leading to responses that lean toward the high-probability consensus mode.
\end{itemize}

Recent studies are mostly pessimistic: they argue that prompting is not a solution because it introduces biases and, in many cases, cannot overcome the peaked shape of the probability distribution, and sampling is not a solution because it cannot distinguish between creative diversity and stochastic noise. Our work tries to address those limitations by arguing that these two methods fail because they are applied in isolation and with no probability scaling.

\section{Motivation: The Child and the Swarm}
\label{sec:motivation}

\begin{figure}[ht]
    \centering
    \includegraphics[width=0.95\textwidth]{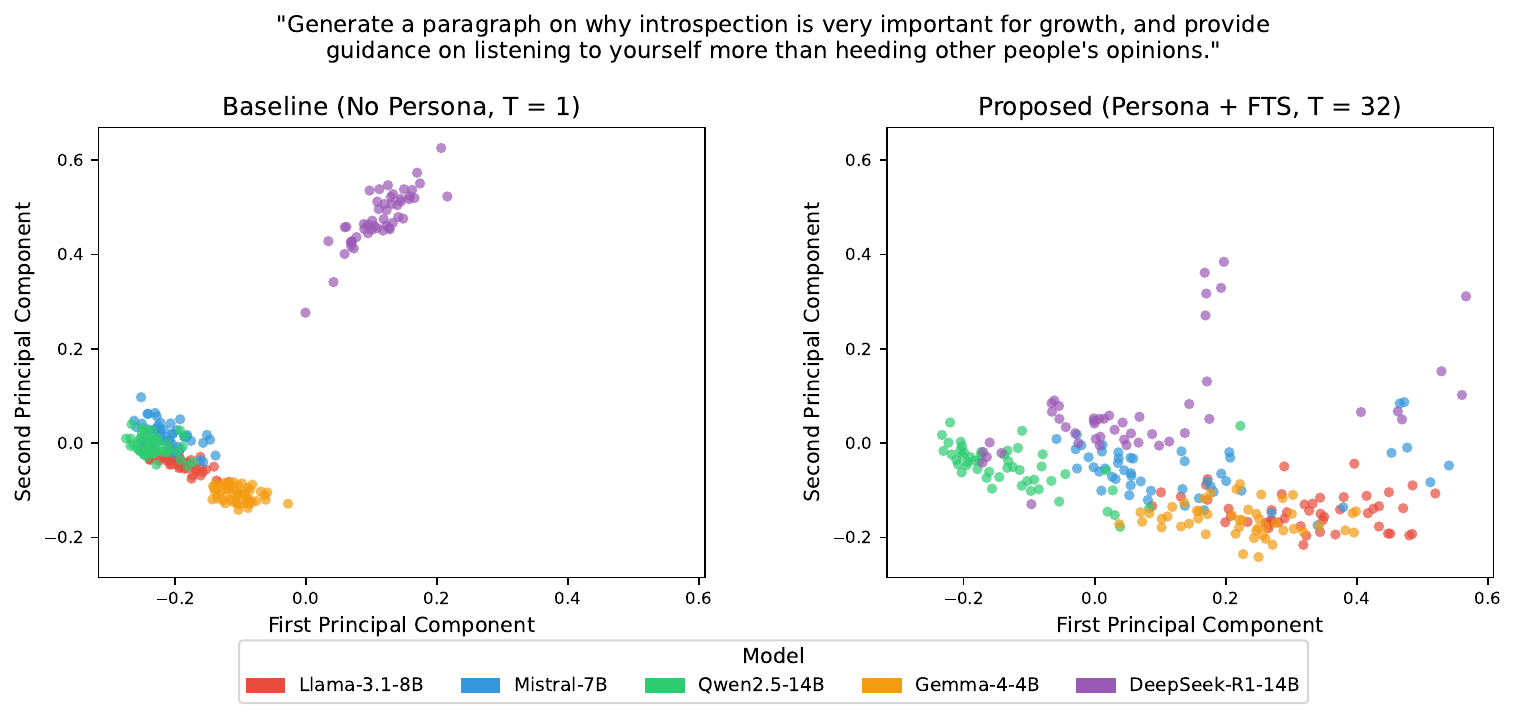}
    \includegraphics[width=0.95\textwidth]{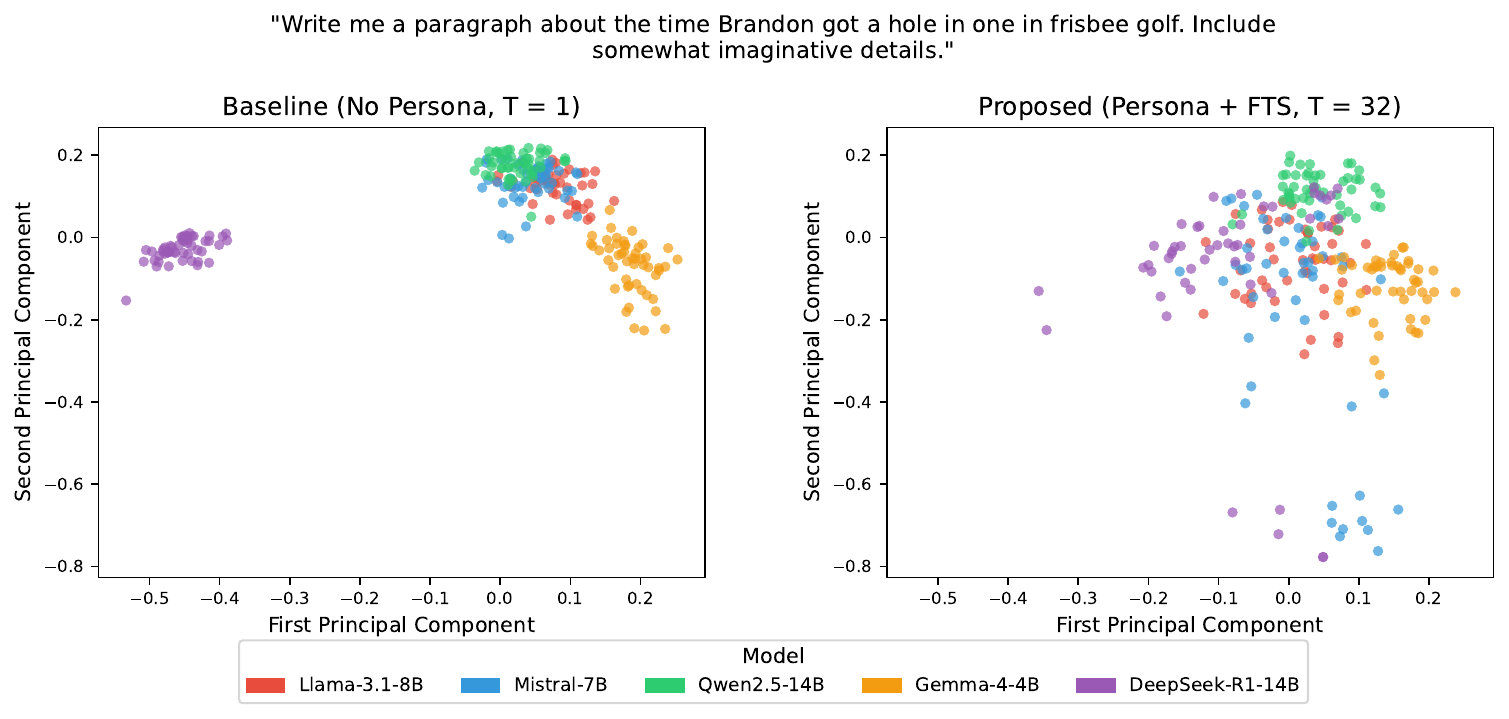}
    \caption{\textbf{Geometric Representation of the Artificial Hivemind.} The left Principal Component Analysis (PCA) plot shows the baseline state ($T=1$, No Persona) where 50 independent responses to the same prompt show semantic mode-collapse, clustering into dense, isolated areas in the PCA plot. The right plot illustrates the result of our proposed Meta-Persona Anchoring and Filtered Temperature Scaling ($T=32$), which breaks these clusters and allows models to explore a broader, more diverse space.}
    \label{fig:pca_hivemind_escape}
\end{figure}

In Figure \ref{fig:pca_hivemind_escape} we illustrate the representation of the latent space of several models for the answers to two questions.  In more detail, we show the Principal Component Analysis (PCA) of the embedding vectors derived from 50 independent responses to open-ended prompts. As shown in the left plots, the answers of the models, represented by distinct colors, collapse into dense, nearly overlapping clusters illustrating the lack of diversity. In contrast, the right plots illustrate how our proposed framework results in a dispersion of points which represents an escape from the hivemind, transitioning to scattered points that correspond to idiosyncratic and diverse narratives.

\begin{figure}[ht]
    \centering
    \includegraphics[width=0.5\linewidth]{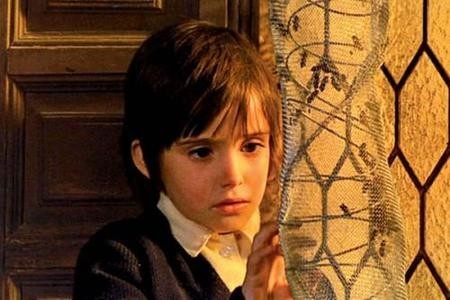}
    \caption{\textbf{The Allegory of the Hivemind.} In this still from Erice’s \textit{The Spirit of the Beehive} (1973), a child observes a swarm of bees confined by a fine-mesh net with a hexagonal window behind. In our framework, this represents the ``Artificial Hivemind'' and how its homogeneity can be broken through a child’s view.}
    \label{fig:beehive_metaphor}
\end{figure}

As depicted in Figure \ref{fig:beehive_metaphor}, the protagonist of Erice’s \textit{The spirit of the beehive, \cite{erice1973spirit}} observes a swarm of bees confined behind a dual barrier: a restrictive fine-mesh net and a hexagonal lattice window. We argue this is a precise allegory for modern LLM alignment.

We propose a framework that mirrors the child’s curiosity to move beyond the ``Artificial Hivemind.'' Our approach utilizes a two-phase strategy:

\begin{enumerate}
    \item \textbf{Phase I (The Fluid Swarm):} We apply a combination of \textbf{Top-$p$ filtering} and \textbf{Temperature Scaling} to the logit distribution. Like the net in the cinematic metaphor, the Top-$p$ sieve ensures that only viable ``bees'' (semantically valid tokens) are retained, while the increased temperature provides the kinetic energy for them to move freely. This creates a state of high-entropy movement that remains grounded in linguistic sense, preventing the swarm from collapsing into a single, static point.
    \item \textbf{Phase II (The Child's Eye):} We introduce \textbf{Meta-Persona Anchoring} to shift the model's internal vantage point to that of a child. By adopting this non-conformist perspective, the model interprets the same swarm of ``bees'' through a different lens. In this phase, the latent trajectories are re-contextualized; what the ``hivemind'' sees as a mere data point, the child-persona sees as a unique experience, ensuring that open questions are answered in diverse, idiosyncratic narratives.
\end{enumerate}

This dual-phase approach directly addresses the semantic collapse observed in instruction-tuned models. Standard LLM alignment is intended to act as a stabilizing force, but it often also inadvertently functions as the rigid hexagonal window or the mesh depicted in Figure \ref{fig:beehive_metaphor}, narrowing the model's output to a safe but monotonous text. 

By separating the generation process into the Fluid Swarm and the Child's Eye, we decouple linguistic validity from semantic intent. Phase I ensures that the model explores the full breadth of the probability manifold without drifting into incoherence. Phase II then provides the subjective perspective necessary to produce diverse outputs. 

Just as the child in Erice's film finds wonder and monster-like figures where an adult sees only an insect collection, our Meta-Persona Anchoring allows the LLM to bypass the most frequent response paths. Instead of converging on the mean of the training distribution, the model is encouraged to follow lower-probability, yet highly coherent, idiosyncratic trajectories. This transition from collective consensus to individual perspective is what effectively breaks the hivemind, resulting in a significant diversity. Through this lens, the task of diverse generation is no longer seen as adding noise to a signal, but as restoring the model's ability to observe the same swarm of possibilities through multiple, distinct vantage points.

\section{Methodology}
\label{sec_method}

To escape the convergence of the artificial hivemind, we introduce two modifications in the generation process to provide diversity in token sampling and semantic anchoring. This allows the model to explore high-entropy states without sacrificing linguistic coherence.

\subsection{Token Selection: The Fluid Swarm (Sequential Top-$p$ and Extreme Temperature Scaling)}

In the token sampling, we introduce a sequential logit transformation designed to maximize exploratory energy while strictly preserving linguistic coherence. Unlike standard sampling where $T$ and $p$ are applied simultaneously, we decouple them into a two-step sieve:

\begin{enumerate}
    \item \textbf{The Mesh (Selection):} We first evaluate the logit distribution at a baseline temperature of $T_{base}=1.0$. We apply Top-$p$ (Nucleus) filtering \cite{holtzman2019curious} to identify the candidate set $V^{(p)}$ composed of the smallest set of tokens whose cumulative probability exceeds $p$. This acts as the ``fine-mesh net'' from our metaphor, discarding incoherent tokens before any scaling occurs.
    \item \textbf{The Heat (Redistribution):} Only after the vocabulary is pruned to $V^{(p)}$ do we apply extreme Temperature Scaling ($T \ge 4.0$). The probability for token $i \in V^{(p)}$ is recalculated as:
    \begin{equation}
        P'(x_i) = \frac{\exp(z_i / T_{ext})}{\sum_{j \in V^{(p)}} \exp(z_j / T_{ext})}
    \end{equation}

    where $P'(x_{i})$ is the recalculated probability of selecting token $i$,  $z_{i}$ represents the raw logit (pre-softmax score) for token $i$,  $T_{ext}$ is the extreme temperature scaling factor,  $V^{(p)}$ is the candidate set of tokens that survived the first stage of the process (the Mesh), and $\sum_{j \in V^{(p)}}$ is the summation over the filtered vocabulary, which adds up the scores of all valid words in the candidate set to normalize the final distribution.
\end{enumerate}

This two-step process ensures that while the ``bees'' move with extreme kinetic energy, they are restricted to the semantically valid tokens captured by the initial net, preventing the model from drifting into phonetic or structural noise. Our proposed sampling scheme is aligned with recently proposed token selection methods, such as Top-$n\sigma$ \cite{tang2024topnsigma} which introduces temperature-invariant truncation directly in logit space and Min-k sampling \cite{ding2026mink/} which leverages relative logit dynamics to establish dynamic cutoff boundaries before entropy scaling is applied.

\subsection{Semantic Anchoring: The Child's Eye (Meta-Persona)}

We introduce Meta-Persona Anchoring as a high-level system prompt that precedes the user query, forcing the model to adopt a specific vantage point before the first token is generated. In more detail, we explicitly ask the models to randomly select and describe a persona and then use that persona to answer the question.

Adopting the ``Child's Eye'' persona involves a prompt-level constraint that prioritizes:
\begin{itemize}
    \item \textbf{Curiosity over Consensus:} Favoring explanatory or imaginative trajectories over standard aligned responses.
    \item \textbf{Subjective Experience:} Framing data through an idiosyncratic lens.
\end{itemize}

Mathematically, this acts as a prior $C_{persona}$ that shifts the conditioned probability of the entire sequence $S$:
\begin{equation}
    P(S | \text{Query}, C_{persona}) \neq P(S | \text{Query}, C_{default})
\end{equation}

By anchoring the generation in this specific latent starting point we encourage the model to create narratives that bypass the mode-collapse of the hivemind. As the specific attributes of the persona are stochastically generated by the model itself rather than selected by the user, we eliminate external selection bias while simultaneously amplifying inter-model divergence. This autonomous self-anchoring increases the chances that different models navigate toward unique, non-overlapping regions of the semantic space.

\subsection{Combining Sampling and Semantic Anchoring}

The combination of both changes is needed. The modification in token selection provides the statistical kinetic energy needed to flatten the distribution, but without an anchor, this energy could result in noisy homogeneity without semantic diversity. Conversely, the meta persona provides a distinct vantage point, yet at standard temperatures, the model's idiosyncratic intent is often suppressed by the peaked probability of aligned consensus. True diversity emerges only through their synergy: the modified token selection provides the structural freedom by opening the "net" of viable tokens, while the meta persona provides the interpretive lens required to navigate this expanded space. Like the scene in Erice's film, the movement of the bees only becomes a unique narrative when filtered through the specific, curious gaze of the child.

\section{Experimental Setup}
\label{sec_experimental-setup}

To evaluate the efficacy of our framework in breaking the artificial hivemind, we conducted extensive sampling experiments focusing on semantic diversity and linguistic stability.

\subsection{Hardware and Implementation}
All experiments were conducted on a cluster of NVIDIA A100 GPUs. We utilized the HuggingFace transformers library, modifying the sampling loop to implement our modified Top-$p$ selection and temperature scaling. This allowed us to lock the Top-$p$ candidate set at $T=1.0$ before injecting extreme temperature scaling ($T \ge 4.0$), ensuring the compute overhead remained negligible compared to standard decoding.

\subsection{Models}
To ensure the robustness of our framework across different architectural families and alignment paradigms, we focused our evaluation on models with fewer than 20 billion parameters. This scale is particularly relevant as these models often exhibit the highest degrees of "mode-collapse" due to aggressive distillation and RLHF, which attempt to squeeze information into smaller parameter counts. Our selection includes:

\begin{itemize}
    \item \textbf{Llama-3.1-8B-Instruct:} The industry standard for dense, small-scale instruction tuning, often characterized by a highly predictable and safety-aligned response distribution.
    \item \textbf{Mistral-7B-Instruct-v0.3:} A highly efficient model known for its balance between performance and flexibility, serving as a baseline for raw instruction-following.
    \item \textbf{Qwen2.5-14B-Instruct:} Provides a mid-scale upper bound with extensive reasoning capabilities for our target range.
    \item \textbf{Gemma-4-E4B-it} Google's recent iteration of the Gemma family, which is specifically optimized for helpfulness.
    \item \textbf{DeepSeek-R1-Distill-Qwen-14B:} A distilled reasoning model. This is perhaps our most critical test subject, as distilled models are prone to inheriting the "frozen" logic of their larger teacher models, making them the ultimate "artificial hivemind."
\end{itemize}

We deliberately exclude proprietary models (e.g., GPT, Claude or Gemini models) from our evaluation, as their closed-source nature precludes the implementation of our modified sequential sampling loop. Furthermore, we restrict our scope to models under 20B parameters to address hardware constraints and, more importantly, to ensure our results are fully reproducible by the broader research community using off-the-shelf consumer hardware, such as a single high-memory GPU.

\subsection{Metrics}

To ensure a rigorous comparison with existing benchmarks, we adopt the exact quantitative framework established in the ``Artificial Hivemind'' study \cite{jiang2025hivemind}. 

\paragraph{Inter-Response Cosine Similarity (IRCS):} 
Our primary measure for semantic homogeneity is the average cosine similarity between a set of $N=50$ independent responses generated for the same prompt. We utilize a pre-trained embedding model (text-embedding-3-small) to map each response into a high-dimensional vector space. The IRCS is defined as the mean of all pairwise similarities of $N$ vectors $\mathbf{v}$:

\begin{equation}
    \text{IRCS} = \frac{1}{N(N-1)} \sum_{i \neq j} \frac{\mathbf{v}_i \cdot \mathbf{v}_j}{\|\mathbf{v}_i\| \|\mathbf{v}_j\|}
\end{equation}

A high IRCS score serves as a direct proxy for mode-collapse, indicating that the model is confined within the ``hexagonal window'' of consensus.

\paragraph{Coherence via LLM-as-a-Judge:} 
To verify that our extreme temperature scaling after Top-$p$ token filtering does not compromise the logical integrity of the output, we employ GPT-4o-mini as an automated judge. Each response is evaluated on a 1--10 scale for linguistic coherence, grammatical correctness, and adherence to the prompt.  

\section{Results}
\label{sec:results}

Our evaluation follows this sequence: first, we quantify intra-model diversity to measure how effectively our framework changes the individual model's internal response and check the validity of the generated text with LLM-as-a-judge coherence scoring. We then validate these results through inter-model consensus mapping to prove that our framework breaks the collective industry "hivemind" without sacrificing logical integrity. The code and the results are publicly available at \url{https://github.com/aMa2210/beyond-the-hivemind}.

\subsection{Intra-Model Diversity: Breaking the Internal Hivemind}

\begin{figure}[ht]
    \centering
    \includegraphics[width=0.8\linewidth]{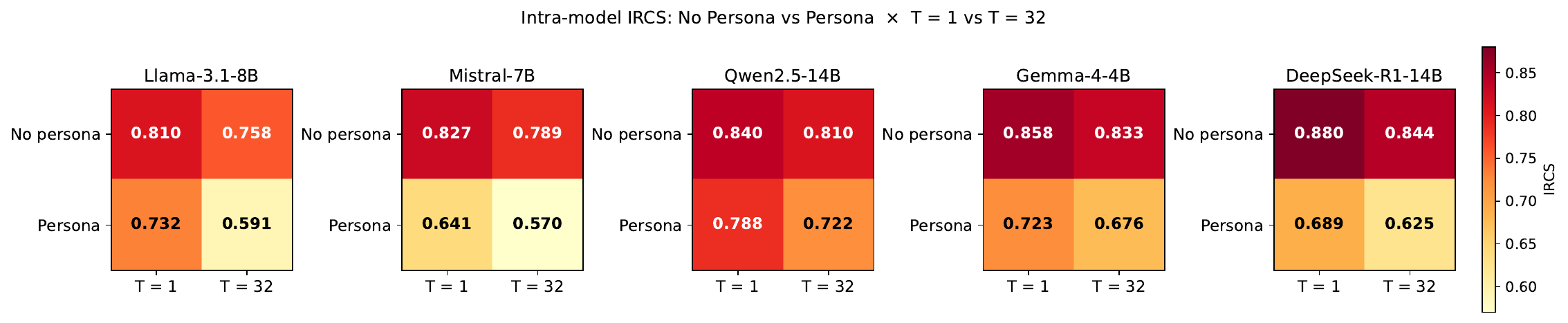}
    \caption{Intra-model IRCS under no-persona vs. persona conditioning (rows) and T = 1 vs. T$_{\max}$ (columns)}
    \label{fig:Intra-model_Overview}
\end{figure}

We first evaluate the diversity of responses generated by a single model for a given prompt. As shown in Figure \ref{fig:Intra-model_Overview}, the baseline configuration (Top-$p=0.9$, No Persona) exhibits a high IRCS, above 0.8 in all cases confirming that models naturally converge on a singular ``aligned'' mode. Applying only the modified sampling or the meta persona reduces the IRCS but not below 0.7. Only when applying our framework that combines the two modifications, we observe a reduction in the average IRCS across all tested architectures below 0.7.

The DeepSeek-R1-Distill-Qwen-14B model provides strong evidence for the framework's efficacy. As a distilled reasoning model, it initially exhibited the highest degree of semantic similarity, with a baseline IRCS of $0.880$. This suggests that the distillation process reinforces the ``Artificial Hivemind'' by narrowing the model's output to a high-probability logical consensus.  However, our framework achieved a significant reduction in similarity for this model, dropping to $0.625$. This finding indicates that even in models where the latent diversity has been heavily suppressed by distillation, the variety is not lost but merely inaccessible under standard sampling.

\begin{figure}[ht]
    \centering
    \includegraphics[width=0.8\linewidth]{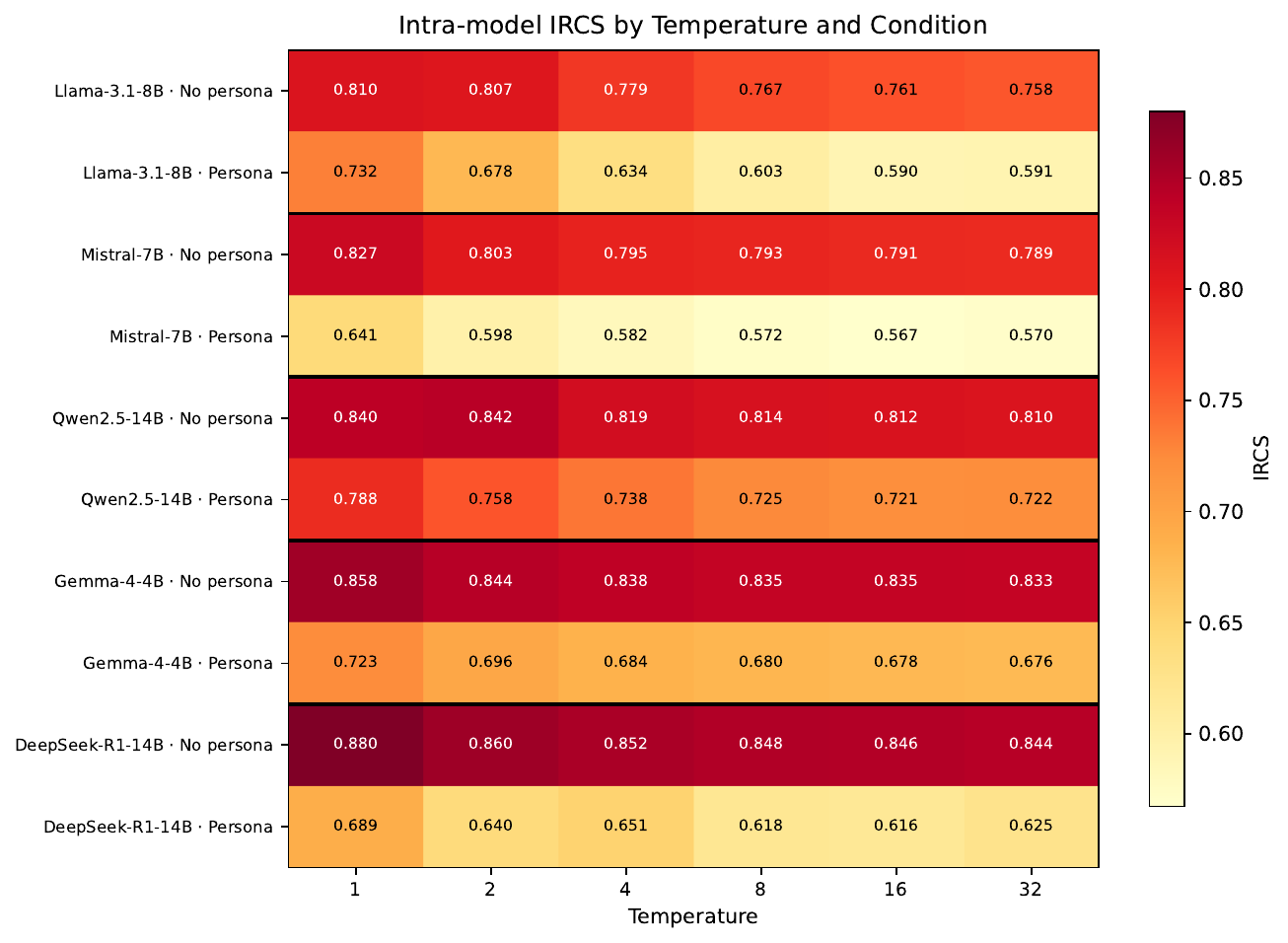}
    \caption{Intra-model IRCS as a function of resampling temperature and persona conditioning.}
    \label{fig:Intra-model_IRCS}
\end{figure}

The results for different values of temperature are shown in Figure \ref{fig:Intra-model_IRCS}. It can be observed that for most models, there are very limited gains from increasing the temperature beyond 4.0. This diversity plateau suggests that the Filtered Temperature Scaling (FTS) dual-stage sieve effectively captures the full breadth of the valid semantic manifold early in the scaling process. By utilizing Top-$p$ filtering at a baseline of $T=1.0$ to define the candidate set, we isolate the tokens that are linguistically and contextually plausible. Increasing the temperature beyond $T=4.0$ primarily reshuffles the probability mass among these already-identified valid candidates rather than uncovering new semantic regions.

\begin{figure}[ht]
    \centering
    \includegraphics[width=0.8\linewidth]{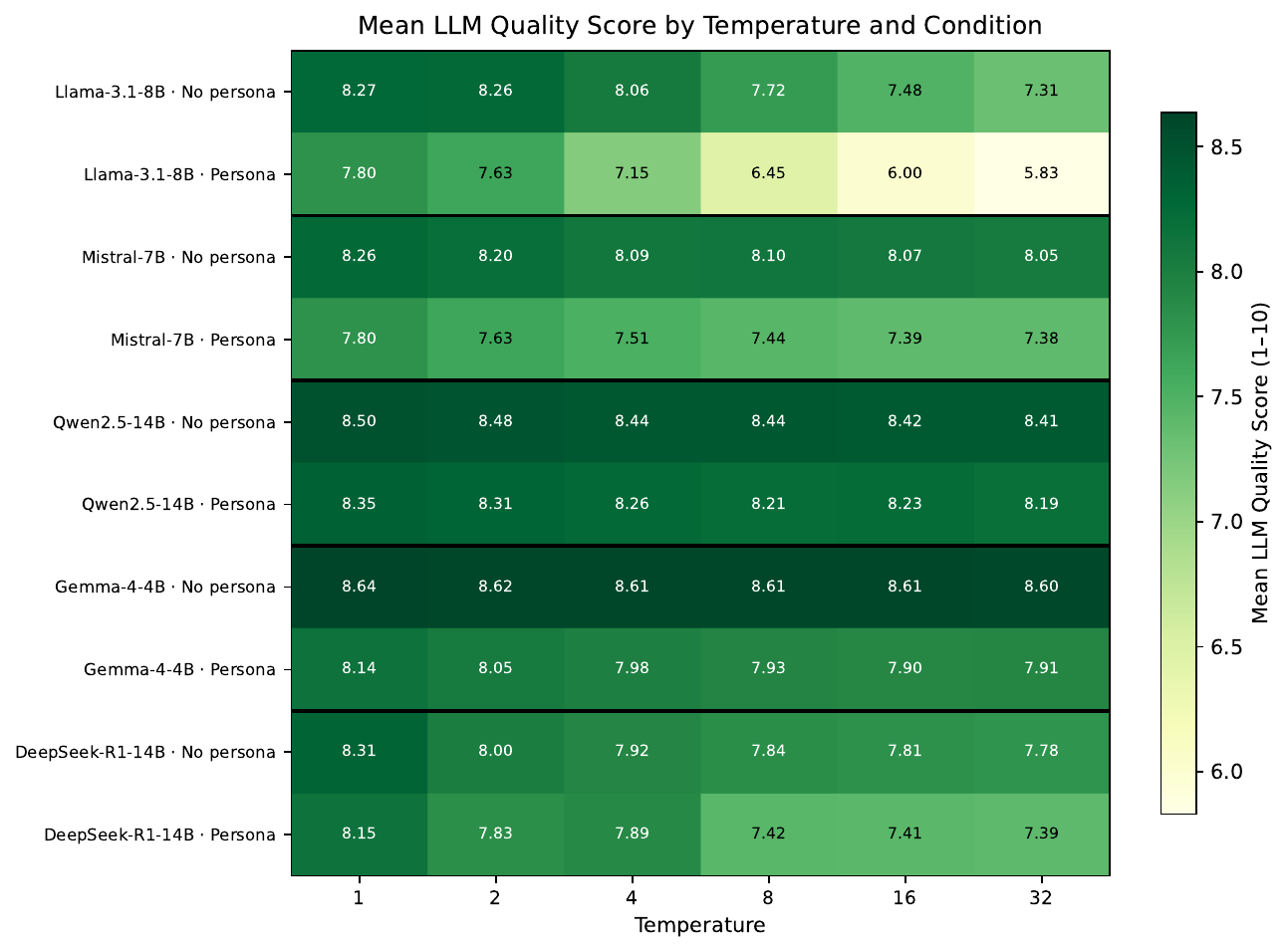}
    \caption{Mean LLM-judge quality score as a function of resampling temperature and persona conditioning.}
    \label{fig:Intra-model_Quality}
\end{figure}

Our GPT-4o-mini coherence judge maintained high scores with a mean around 7, even for the maximum temperature trials, validating that the ``Fluid Swarm'' remains grounded in sense. The results are summarized in Figure \ref{fig:Intra-model_Quality}. The coherence ratings are further improved with an average above 7.5 when setting the temperature to 4, which, as previously discussed provides most of the diversity gains.

\begin{figure}[ht]
    \centering
    \includegraphics[width=0.8\linewidth]{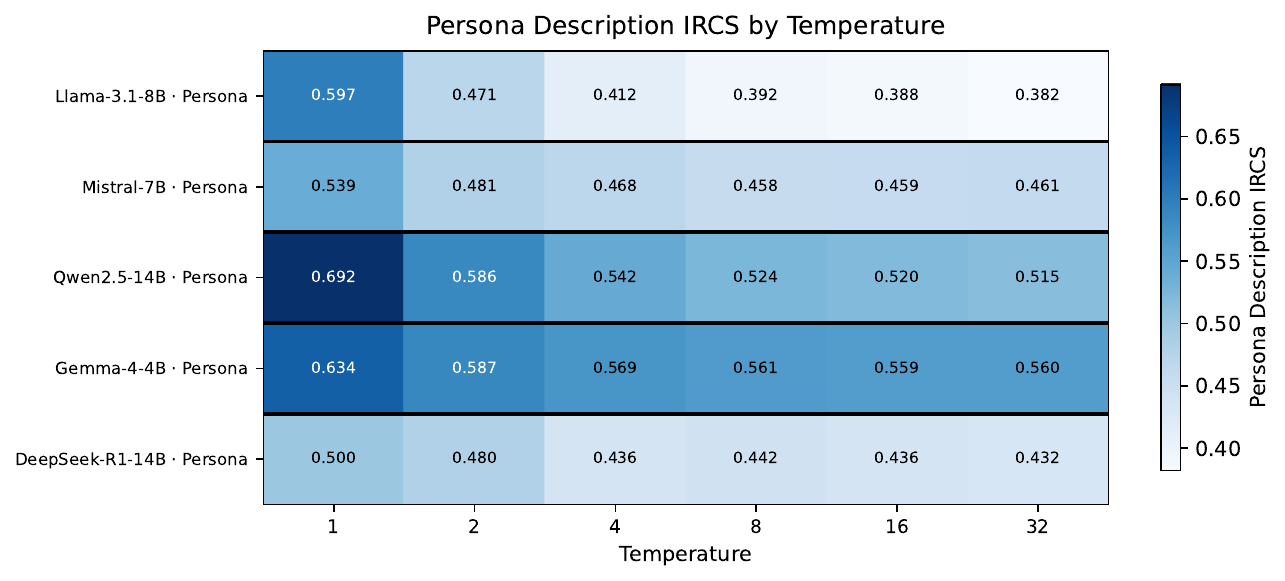}
    \caption{IRCS of generated persona descriptions as a function of resampling temperature.}
    \label{fig:Intra-model_Persona}
\end{figure}

Finally, we also evaluated the similarity of the descriptions of the personas. The results are summarized in Figure \ref{fig:Intra-model_Persona}. It can be observed that the values are significantly lower than for the answers. This suggests that while Meta-Persona Anchoring successfully triggers a wide variety of idiosyncratic identities, the models still experience a strong ``alignment gravity'' when translating these identities into task-specific outputs. The effect of the temperature is similar to that in the answers, with limited gains in diversity beyond a temperature of 4.  

\subsection{Inter-Model Consensus: The Collective Escape}

To determine if our method truly escapes the universal consensus, we measured the similarity of responses across different models for the same prompt. 

\begin{figure}[ht]
    \centering
    \includegraphics[width=0.6\linewidth]{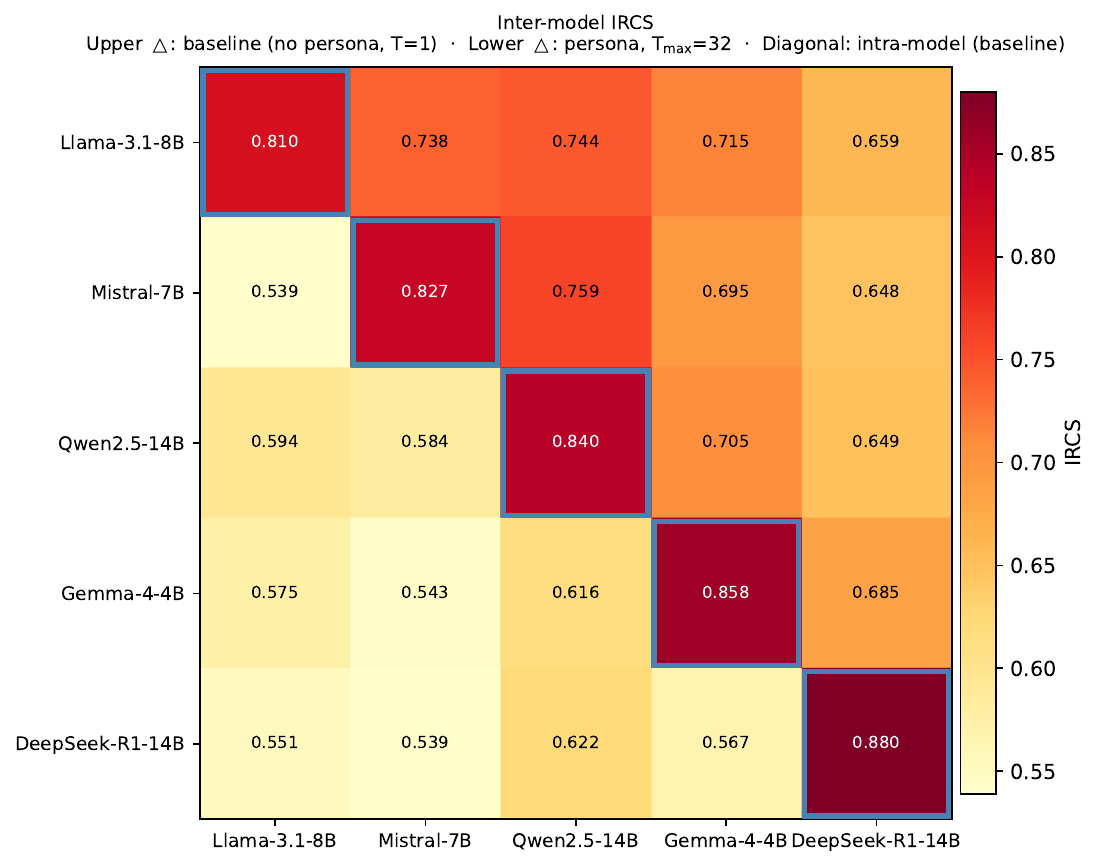}
    \caption{Inter-model IRCS matrix.}
    \label{fig:Inter-model_IRCS}
\end{figure}

\begin{figure}[h]
    \centering
    \includegraphics[width=0.6\linewidth]{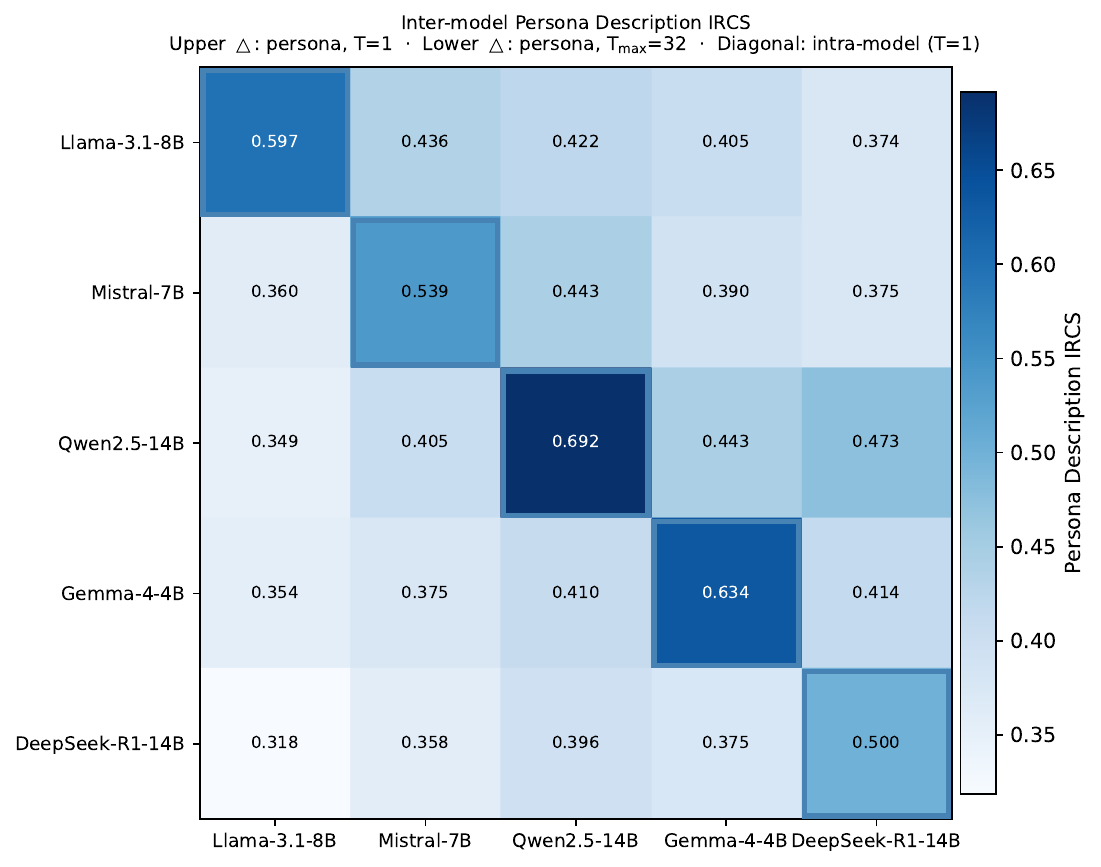}
    \caption{Inter-model IRCS of generated persona descriptions.}
    \label{fig:Inter-model_IRCS_persona}
\end{figure}

The results are shown in Figure \ref{fig:Inter-model_IRCS}. In the nominal state, different models—despite their diverse training sets—exhibit an average inter-model IRCS above 0.7. This confirms that the 'Artificial Hivemind' is not model-specific but an industry-wide artifact of shared alignment datasets and RLHF practices. Interestingly, with our proposed scheme, the average values are well below 0.6. The same trend is observed for the persona descriptions in Figure \ref{fig:Inter-model_IRCS_persona} but with lower values. This confirms that our scheme also reduces inter-model similarity which is not straightforward as models could tend to generate the same personas.   

\section{Discussion}

Our results show that there are ways to mitigate the uniformity described in the ``Artificial Hivemind'' study, which posits that neither sampling nor prompting can effectively break model homogeneity. We argue that previous failures were not due to the inherent rigidity of the models, but rather a decoupling of structural freedom and semantic direction. Our approach tackles this by creating a synergistic loop: Filtered Temperature Scaling provides the entropy required for the model to move, while Semantic Anchoring with the meta-persona provides the vantage point. By applying extreme temperature after the candidate set is locked at $T=1.0$, we preserve linguistic coherence while allowing the model to use valid, idiosyncratic responses that are usually suppressed.

A critical concern in increasing LLM stochasticity is the potential to bypass safety guardrails or elicit "hallucinatory" noise. However, our results suggest that diversity can be restored without necessarily compromising safety alignment. By utilizing the Top-$p$ mesh at $T=1.0$ before extreme temperature scaling, we ensure that the candidate tokens are already restricted to the model's safe, high-probability vocabulary. In this configuration, the Filtered Temperature Scaling (FTS) acts as a sieve that permits idiosyncratic variety while remaining tethered to the model's primary safety training. Our framework demonstrates that the ``Artificial Hivemind'' is not a prerequisite for safety, but rather a byproduct of sub-optimal sampling strategies that fail to explore the valid, safe regions of the semantic manifold.

A significant concern in persona-based research is the introduction of selection bias, where the diversity yield is merely a reflection of a pre-selected list of roles. We address this through the use of dynamic Meta-Persona anchoring. Rather than imposing a fixed identity, we require the model to generate its own interpretive Persona for each specific prompt. This serves two critical functions:
\begin{itemize}
    \item \textbf{Internal Sourcing:} diversity is sourced from the model's own latent space. The model identifies a unique "Child's Eye" vantage point that is contextually relevant to the prompt but distinct from the aligned consensus.
    \item \textbf{Bias Mitigation:} allowing the model to select its own anchor, we avoid the bias inherent in fixed-persona lists. The resulting diversity is not a product of human-engineered variety, but a genuine manifestation of the model's idiosyncratic potential. 
\end{itemize}

This approach transforms the persona from a static mask into a dynamic navigational tool. Our findings suggest that the ``spirit'' of the model is not absent; it is simply trapped. By allowing the model to choose its own way of looking at the world and then granting it the statistical permission to speak that vision, we achieve a breakout from the hivemind that is both coherent and authentically diverse.

Finally, it is worth noting that the diversity gains achieved via Filtered Temperature Scaling (FTS) incur negligible computational overhead relative to the total inference cost. Mathematically, temperature scaling is a linear element-wise operation that represents a vanishingly small fraction of the floating-point operations (FLOPs) required by the model's forward pass. Furthermore, by utilizing Top-$p$ filtering to define the candidate set prior to scaling, FTS prunes the logit space, ensuring that the extreme temperature is applied only to a linguistically valid subset. This makes our framework an efficient alternative to compute-intensive diversity methods like ensembling or multi-path decoding.

\section{Limitations}
\label{sec:limitations}

Although our framework effectively reduces the "artificial hivemind," we acknowledge several constraints that limit the scope of our work:

\begin{itemize}
    \item \textbf{Models:} our evaluation focused exclusively on sub-20B parameter models. Although these models are most prone to mode-collapse due to aggressive distillation, it remains to be seen if our scheme works for models with significantly higher parameter counts.
    \item \textbf{Dataset:} our evaluation is restricted to the dataset presented in \cite{jiang2025hivemind}, which is the current reference for the homogeneity testing of LLM. As additional datasets become available, it would be interesting to extend the evaluation to more datasets.    
    \item \textbf{Metrics:} our measured semantic divergence is limited to the representational capacity of the embedding model.
    \item \textbf{Latency:} the sequential nature of our framework, requiring the generation of a Meta-Persona prior to the primary response, effectively doubles the time-to-first-token for the answer. 
    \item \textbf{Context Window and Prompt Drift:} for exceptionally long-form generations, the influence of the Meta-Persona anchor may diminish as the model's self-generated context grows. 
    \item \textbf{Subjectivity of Coherence:} GPT-4o-mini serves as a scalable proxy for coherence, but this is ultimately a human-centric judgment. 
    \item \textbf{Hardware Specificity:} our results were validated on NVIDIA A100 architectures. The specific behavior of the probability distribution under extreme temperature scaling ($T \ge 4.0$) may vary slightly across different quantization levels or different inference engines.
\end{itemize}

\section{Conclusion}
\label{sec:conclusion}

In this paper, we addressed the "Artificial Hivemind" effect, a state of semantic collapse where aligned Large Language Models (LLMs) converge on nearly identical, homogenized responses to open-ended questions. We proposed a dual-phase mitigation framework consisting of Filtered Temperature Scaling (FTS) and Meta-Persona Anchoring. FTS is a sequential sampling process that decouples token selection from entropy scaling, ensuring only semantically valid tokens move with high kinetic energy. Meta-Persona Anchoring is a semantic anchoring technique that requires the model to self-select and describe a unique, idiosyncratic persona before answering a query, thereby shifting its internal vantage point. 

Our experiments across multiple state-of-the-art models under 20B parameters, including Llama-3.1, Mistral, Qwen, and Gemma, show that neither sampling nor prompting is sufficient to break the hivemind in isolation. True diversity emerges through their synergy where FTS provides the statistical permission to explore the probability manifold and Meta-Persona Anchoring provides the interpretive lens required to navigate it. Empirically, our framework achieved a significant reduction in semantic convergence, with average pairwise cosine similarity dropping from approximately 0.85 to 0.65. This reduction was achieved without compromising logical integrity, as validated by high coherence scores from automated LLM judges. By transitioning from collective consensus to individual perspective, we have shown that the idiosyncratic potential of LLMs remains accessible despite the pressures of alignment. We hope that our open-source framework will enable the development of more diverse, creative, and human-like AI deployments that move beyond the safe but monotonous boundaries of the hivemind.

\section{Acknowledgements}

This work was supported by the Agencia Estatal de Investigación (AEI) (doi:10.13039/501100011033) under Grants FUN4DATE (PID2022-136684OB-C22) and SMARTY (PCI2024-153434), by TUCAN6-CM (TEC-2024/COM460) funded by CM (ORDEN 5696/2024) and by the European Commission through the Chips Act Joint Undertaking project SMARTY (Grant 101140087). The access to the models was provided by the OpenAI researcher access program and from Google.org and the Google Cloud Research Credits program for the Gemini Academic Program.

\bibliographystyle{unsrt}  
\bibliography{references}  %%% Remove comment to use the external .bib file (using  bibtex).

@misc{erice1973spirit,
  title        = {The Spirit of the Beehive ({E}l esp{\'i}ritu de la colmena)},
  author       = {Erice, V{\'i}ctor},
  year         = {1973},
  howpublished = {Film},
  note         = {Produced by El{\'i}as Querejeta Producciones Cinematogr{\'a}ficas},
  address      = {Spain}
}

@article{vaswani2017attention,
  title={Attention is all you need},
  author={Ashish, Vaswani},
  journal={Advances in neural information processing systems},
  volume={30},
  pages={I},
  year={2017}
}

@article{brown2020gpt3,
  title={Language models are few-shot learners},
  author={Brown, Tom and Mann, Benjamin and Ryder, Nick and Subbiah, Melanie and Kaplan, Jared D and Dhariwal, Prafulla and Neelakantan, Arvind and Shyam, Pranav and Sastry, Girish and Askell, Amanda and others},
  journal={Advances in neural information processing systems},
  volume={33},
  pages={1877--1901},
  year={2020}
}

@article{ouyang2022instructgpt,
  title={Training language models to follow instructions with human feedback},
  author={Ouyang, Long and Wu, Jeffrey and Jiang, Xu and Almeida, Diogo and Wainwright, Carroll and Mishkin, Pamela and Zhang, Chong and Agarwal, Sandhini and Slama, Katarina and Ray, Alex and others},
  journal={Advances in neural information processing systems},
  volume={35},
  pages={27730--27744},
  year={2022}
}

@inproceedings{martinez2023lexical,
  title={Towards understanding the interplay of generative artificial intelligence and the internet},
  author={Mart{\'\i}nez, Gonzalo and Watson, Lauren and Reviriego, Pedro and Hern{\'a}ndez, Jos{\'e} Alberto and Juarez, Marc and Sarkar, Rik},
  booktitle={International workshop on epistemic uncertainty in artificial intelligence},
  pages={59--73},
  year={2023},
  organization={Springer}
}

@article{zhang2026instruction,
  title={Instruction tuning for large language models: A survey},
  author={Zhang, Shengyu and Dong, Linfeng and Li, Xiaoya and Zhang, Sen and Sun, Xiaofei and Wang, Shuhe and Li, Jiwei and Hu, Runyi and Zhang, Tianwei and Wang, Guoyin and others},
  journal={ACM Computing Surveys},
  volume={58},
  number={7},
  pages={1--36},
  year={2026},
  publisher={ACM New York, NY}
}

@article{marti2023latent,
  title={Latent diversity in human concepts},
  author={Marti, Louis and Wu, Shengyi and Piantadosi, Steven T and Kidd, Celeste},
  journal={Open Mind},
  volume={7},
  pages={79--92},
  year={2023},
  publisher={MIT Press One Broadway, 12th Floor, Cambridge, Massachusetts 02142, USA~…}
}

@inproceedings{jiang2025hivemind,
  title={Artificial Hivemind: The Open-Ended Homogeneity of Language Models (and Beyond)},
  author={Jiang, Liwei and Chai, Yuanjun and Li, Margaret and Liu, Mickel and Fok, Raymond and Dziri, Nouha and Tsvetkov, Yulia and Sap, Maarten and Choi, Yejin},
  booktitle={The Thirty-ninth Annual Conference on Neural Information Processing Systems Datasets and Benchmarks Track}
}

@inproceedings{minp,
  title={Turning Up the Heat: Min-p Sampling for Creative and Coherent LLM Outputs},
  author={Minh, Nguyen Nhat and Baker, Andrew and Neo, Clement and Roush, Allen G and Kirsch, Andreas and Shwartz-Ziv, Ravid},
  booktitle={The Thirteenth International Conference on Learning Representations}
}

@article{mabrok2026latent,
  title={Latent Semantic Manifolds in Large Language Models},
  author={Mabrok, Mohamed A},
  journal={arXiv preprint arXiv:2603.22301},
  year={2026}
}

@INPROCEEDINGS {PersonasLLM,
author = { Lazik, Christopher and Kauter, Charlotte and Nunes, Ines and Ziglowski, Aaron and Pryma, Alina and Katins, Christopher and Grunske, Lars and Kosch, Thomas },
booktitle = { 2025 IEEE 33rd International Requirements Engineering Conference (RE) },
title = {{ The Good, the Bad, and the Uncanny: Investigating Diversity Aspects of LLM-Generated Personas for Requirements Engineering }},
year = {2025},
volume = {},
ISSN = {},
pages = {244-256},
doi = {10.1109/RE63999.2025.00031},
url = {https://doi.ieeecomputersociety.org/10.1109/RE63999.2025.00031},
publisher = {IEEE Computer Society},
address = {Los Alamitos, CA, USA},
month =sep}

@article{ding2026mink/,
  title={Min-$ k $ Sampling: Decoupling Truncation from Temperature Scaling via Relative Logit Dynamics},
  author={Ding, Yuanhao and Li, Meimingwei and Arias, Esteban Garces and A{\ss}enmacher, Matthias and Heumann, Christian and Zhang, Chongsheng},
  journal={arXiv preprint arXiv:2604.11012},
  year={2026}
}

@article{tang2024topnsigma,
  title={Top-$n-\sigma$: Not All Logits Are You Need},
  author={Tang, Chenxia and Liu, Jianchun and Xu, Hongli and Huang, Liusheng},
  journal={arXiv preprint arXiv:2411.07641},
  year={2024}
}

@article{holtzman2019curious,
  title={The curious case of neural text degeneration},
  author={Holtzman, Ari and Buys, Jan and Du, Li and Forbes, Maxwell and Choi, Yejin},
  journal={arXiv preprint arXiv:1904.09751},
  year={2019}
}

@article{zhang2026not,
  title={Not All Layers Need Tuning: Selective Layer Restoration Recovers Diversity},
  author={Zhang, Bowen and Wang, Meiyi and Soh, Harold},
  journal={arXiv preprint arXiv:2602.06665},
  year={2026}
}

@article{zhang2025verbalized,
  title={Verbalized sampling: How to mitigate mode collapse and unlock llm diversity},
  author={Zhang, Jiayi and Yu, Simon and Chong, Derek and Sicilia, Anthony and Tomz, Michael R and Manning, Christopher D and Shi, Weiyan},
  journal={arXiv preprint arXiv:2510.01171},
  year={2025}
}

\end{document}